\def\year{2020}\relax
\documentclass[letterpaper]{article} 
\usepackage{aaai20}  
\usepackage{times}  
\usepackage{helvet} 
\usepackage{courier}  
\usepackage[hyphens]{url}  
\usepackage{graphicx} 
\usepackage{graphicx}  
\usepackage{graphicx}
\usepackage{clrscode3e}
\usepackage{amsthm}
\usepackage{amssymb}
\usepackage{amsmath}
\usepackage{multicol}
\usepackage{subcaption}

\usepackage{listings}
\theoremstyle{plain}\newtheorem{thm}{Theorem}
\theoremstyle{definition}
\theoremstyle{plain}
\theoremstyle{plain}

\usepackage{todonotes}

\newcommand{\inp}[1]{#1.\id{input}}
\newcommand{\domain}[1]{#1.\id{domain}}
\newcommand{\out}[1]{#1.\id{output}}
\newcommand{\graph}[1]{#1.\id{certified}}

\newcommand{\cer}{certified}

\newcommand{\evaluate}[1]{\kw{next}(#1)}
\newcommand{\pluseq}{\mathrel{+}=}

\newcommand{\pre}[1]{\id{pre}(#1)}
\newcommand{\eff}[1]{\id{eff}(#1)}
\newcommand{\pddl}[1]{{\texttt{#1}}} 
\newcommand{\pddlsmall}[1]{{\small \texttt{#1}}} 

\newcommand{\level}[1]{\proc{level}(#1)}

\usepackage{colortbl}
\definecolor{Gray}{gray}{0.85}
\newcolumntype{g}{>{\columncolor{Gray}}c}

\title{PDDLStream: Integrating Symbolic Planners and Blackbox Samplers via Optimistic Adaptive Planning}
\author{
Caelan Reed Garrett, 
Tom\'as Lozano-P\'erez, and 
Leslie Pack Kaelbling\\ 
Computer Science and Artificial Intelligence Laboratory\\ 
Massachusetts Institute of Technology \\
\{caelan, tlp, lpk\}@csail.mit.edu
\thanks{We gratefully acknowledge support from NSF grants 1523767 and 1723381; from AFOSR grant FA9550-17-1-0165; from ONR grant N00014-18-1-2847; from the Honda Research Institute; and from SUTD Temasek Laboratories.  Any opinions, findings, and conclusions expressed in this material are those of the authors and do not necessarily reflect the views of our sponsors.} 
}
\begin{document}

\maketitle

\begin{abstract}
Many planning applications involve complex relationships defined on high-dimensional, continuous variables. For example, robotic manipulation requires planning with kinematic, collision, visibility, and motion constraints involving robot configurations, object poses, and robot trajectories. These constraints typically require specialized procedures to sample satisfying values. We extend PDDL to support a generic, declarative specification for these procedures that treats their implementation as black boxes. 
We provide domain-independent algorithms that reduce PDDLStream problems to a sequence of finite PDDL problems. 
We also introduce an algorithm that dynamically balances exploring new candidate plans and exploiting existing ones.
This enables the algorithm to greedily search the space of parameter bindings to more quickly solve tightly-constrained problems as well as locally optimize to produce low-cost solutions.
We evaluate our algorithms on three simulated robotic planning domains as well as several real-world robotic tasks.
\end{abstract}

\section{Introduction}

Many important planning domains occur in continuous spaces involving complex constraints among variables. 
Consider planning for an 11 degree-of-freedom (DOF) robot tasked with rearranging blocks. 
The robot must find a sequence of \pddlsmall{move}, \pddlsmall{pick}, and \pddlsmall{place} actions involving continuous variables such as robot configurations, robot trajectories, block poses, and block grasps that satisfy complicated kinematic, collision, visibility, and motion constraints, which affect the feasibility of the actions.
Often, special purpose procedures for evaluating and producing satisfying values for these constraints, such as inverse kinematic solvers, collision checkers, and motion planners, are known. 

\begin{figure}[ht]
  \centering
    \includegraphics[width=.22\textwidth]{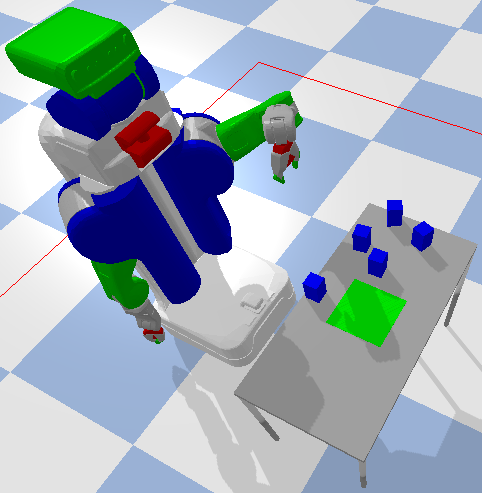}
    \includegraphics[width=.24\textwidth]{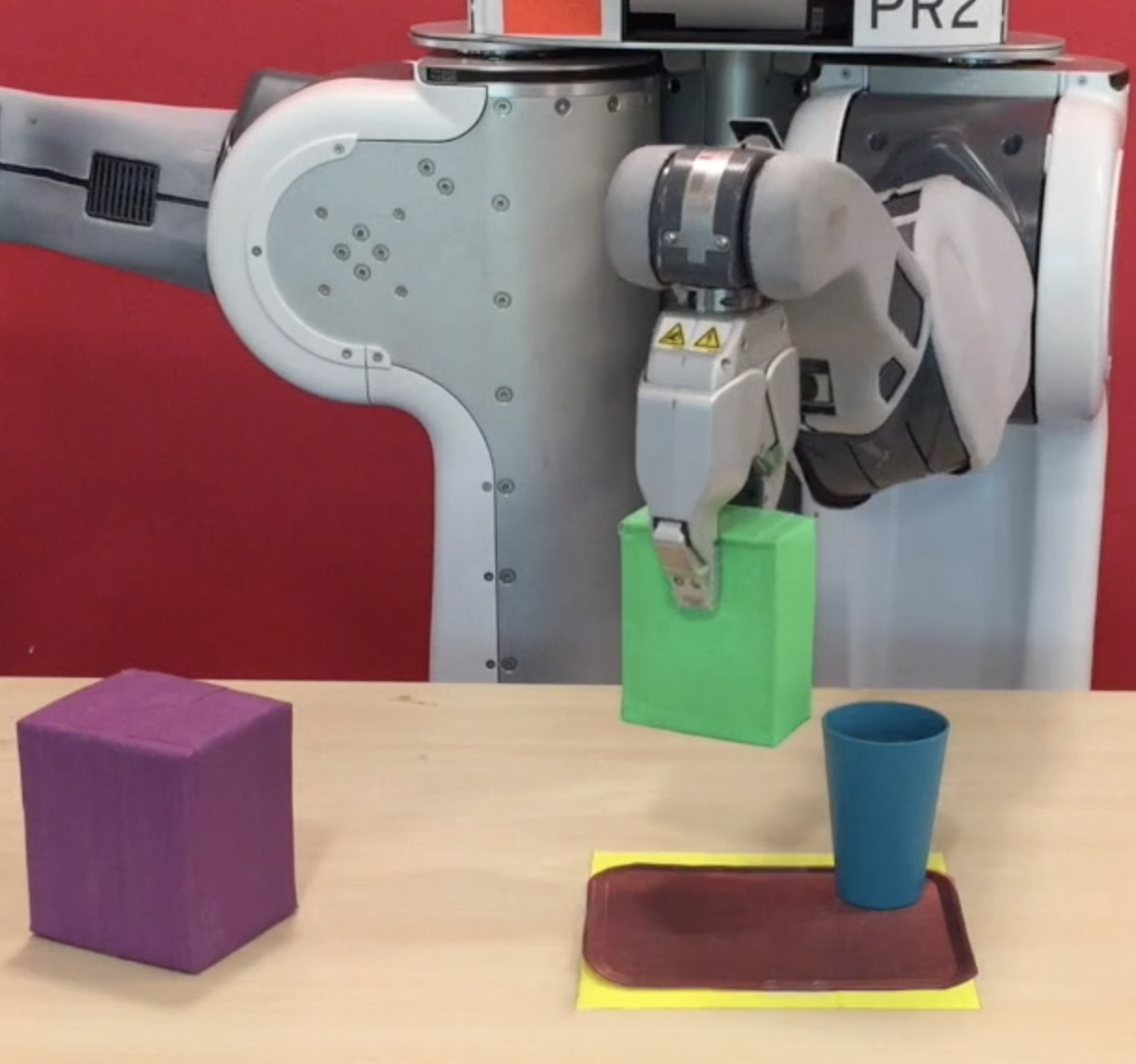}
\caption{Left: {\em Domain 1} (with 5 blocks). 
Right: A real-world robot planning to ``serve a meal" on the brown tray. 
} \label{fig:domain2}
\end{figure}

We propose PDDLStream, a planning language that introduces {\em streams} as an interface for incorporating sampling procedures in Planning Domain Definition Language (PDDL)~\cite{mcdermott1998pddl}.
Streams have both a procedural and declarative component.
The procedural component is a {\em conditional generator}, a function from input values to a possibly infinite sequence of output values. 
Conditional generators construct new values that depend on existing values, such as new robot configurations that satisfy a kinematic constraint with existing poses and grasps. 
The declarative component specifies the facts that these input and output values satisfy.
Streams allow a planner to reason about conditions on the inputs and outputs of a conditional generator while treating its implementation as a black box.



We apply two {\em existing} algorithms~\cite{garrettIJRR2018}  to PDDLStream and introduce two {\em new} PDDLStream algorithms. 
Each algorithm constructs and solves a sequence of finite PDDL problems, {\em any} off-the-shelf PDDL planner to be used as a search subroutine. 
Our {\em Adaptive} algorithm balances the {\em exploration-exploitation trade-off}~\cite{robbins1952some} when deciding whether to search for new {\em optimistic} plans or to continue sampling parameter values for existing ones.
By adaptively balancing the time spent searching versus sampling, {\em Adaptive} is often able to more aggressively find parameter {\em bindings} for existing optimistic plans. 
We experiment in three robotic planning domains (figures~\ref{fig:domain2},~\ref{fig:domain1}, and ~\ref{fig:domain1-plots}) to compare the algorithms.
{\em Adaptive} greatly outperforms the two existing algorithms~\cite{garrettIJRR2018} on constrained and cost-sensitive problems. 
Finally, we apply PDDLStream to a real-world robot to plan for manipulation and kitchen tasks.
 


\section{Related Work}


Several PDDL extensions such as PDDL2.1~\cite{Fox03pddl2.1:an} and PDDL+~\cite{fox2006modelling} support planning with numeric variables that evolve over time. 
Most numeric planners are limited to problems with linear or polynomial dynamics~\cite{hoffmann2003metric,bryce2015smt,cashmore2016compilation}; however, some planners can handle non-polynomial dynamics by discretizing time~\cite{della2009upmurphi,piotrowski2016heuristic}. 
While it may be technically possible to analytically model, for example, collision constraints among 3D meshes using PDDL+, the resulting encoding would be enormous, far exceeding the capabilities of numeric planners.
One approach addresses problems with convex dynamics without discretization~\cite{fernandez2018scottyactivity}; however, it requires a convex decomposition of the robot's configuration space, which is intractable for 3D articulated robots.

Semantic attachments~\cite{dornhege09icaps,dornhege09ssrr,gregory2012planning,hertle2012planning,dornhege2014task,dantam2016tmp}, functions computed by an external module, are an existing method that integrates blackbox procedures and PDDL planners.
Condition-checker modules test Boolean action preconditions, and effect-applicator modules modify numeric state variables. 
Actions must be parameterized by finite types, which restricts the technique to finite action spaces.
In the context of robotics, this restricts the applicability of semantic attachments to domains that are {\em prediscretized}, where a {\em human} specifies a finite set of object poses, object grasps, and robot configurations that can be considered.
Thus, semantic attachments are not sufficient for modeling the domains we consider, where the {\em planner} must produce these continuous values.
In contrast, PDDLStream is able to model domains with infinitely-many action instances.
Finally, semantic attachments are evaluated {\em eagerly} (section~\ref{sec:incremental}) during the forward state-space search as opposed to {\em lazily} (section~\ref{sec:focused}).  
This results in many unneeded module calls and thus poor planner performance when the attachments are computationally expensive. 
Many approaches to robotic task and motion planning have developed strategies for handling continuous spaces that go beyond prediscretization~\cite{HPN,Srivastava14,GarrettIROS15,toussaint2015logic,garrettIJRR2017}. 
However, these approaches are each specialized to a particular class of manipulation problems.
Moreover, they cannot be applied new domains, such as the rovers domain in figure~\ref{fig:domain1}, without substantial engineering effort because they do not offer a modular, domain-agnostic problem description language with clear semantics.

\section{PDDLStream} \label{sec:pddlstream}

We build PDDLStream on PDDL~\cite{mcdermott1998pddl} to enable ease of use for AI practitioners as well as to leverage any PDDL planner, without modification, as a subroutine.
We provide an example PDDLStream specification for a robotic pick-and-place domain in section~\ref{sec:example}.
For clarity of exposition, we formalize STRIPS~\cite{Fikes71} PDDL problems; however, our approach also applies to Action Description Language (ADL)~\cite{pednault1989adl} features such as typing, disjunctions, negative preconditions, existential quantifiers, finite universal quantifiers, conditional effects, and derived predicates.

A {\em predicate} $p$ is a Boolean function.
We treat {\em types} as unary predicates.
An atomic {\em fact} $p(\bar{x})$ is a predicate $p$ evaluated on {\em object} tuple $\bar{x} = \langle x_1, ..., x_k \rangle$ that evaluates to true. 
A {\em literal} is a fact or a negated fact.
A {\em state} ${\cal I}$ is a set of literals.
By the closed world assumption, facts not explicitly specified within a state are false.
An {\em action} $a$ is given by a {\em parameter} tuple $\bar{X} = \langle X_1, ..., X_k \rangle$, a set of literal {\em preconditions} $\pre{a}$ on $\bar{X}$, and a set of literal {\em effects} $\eff{a}$ on $\bar{X}$.
In {\em cost-sensitive} planning, each action may have a nonnegative {\em cost function} $c(\bar{X})$ as an additive cost term.
An {\em action instance} $a(\bar{x})$ is an action $a$ with its parameters $\bar{X}$ replaced with objects $\bar{x}$. 
An action instance $a(\bar{x})$ is {\em applicable} in a state ${\cal I}$ if $(\id{pre}^+(a(\bar{x})) \subseteq {\cal I}) \wedge (\id{pre}^-(a(\bar{x})) \cap {\cal I} = \emptyset)$ where the $+$ and $-$ superscripts designate the positive and negative literals respectively.
The result of {\em applying} an action instance $a(\bar{x})$ to state ${\cal I}$ is a new state $({\cal I} \setminus \id{eff}^-(a(\bar{x}))) \cup \id{eff}^+(a(\bar{x}))$.
To compactly model the domain in section~\ref{sec:example}, we make use of {\em derived predicates} ({\em axioms})~\cite{Fox03pddl2.1:an,thiebaux2005defense}, which are defined by a logical formula on a state.
We treat positive-mentioned instantiated axioms roughly as actions for the purpose of describing the algorithms. 
A STRIPS PDDL {\em problem} $({\cal A}, {\cal I}, {\cal G})$ is given by a set of actions ${\cal A}$, an initial state ${\cal I}$, and a goal set of literals ${\cal G}$.
A {\em plan} $\pi = [a_1(\bar{x}_1), ..., a_k(\bar{x}_k)]$ is a finite sequence of $k$ action instances such that each $a_i(\bar{x}_i)$ is applicable in the $(i-1)$th state resulting from their application.
The {\em preimage} of a consistent plan $\pi$ is the set of facts that must hold to make $\pi$ executable: 
\begin{equation*}
\proc{preimage}(\pi) = \bigcup_{i =1}^k \Big(\pre{a_i(\bar{x}_i)} - \bigcup_{j <i} \eff{a_j(\bar{x}_j)} \Big).
\end{equation*}

\subsection{Streams}


A {\em generator} $g = [\bar{y}_1, \bar{y}_2, ... ]$ is a finite or infinite, enumerable sequence of object tuples $\bar{y}_i$.
Let $\evaluate{g}$ {\em evaluate} the generator and return the subsequent $\bar{y}_i$ in the sequence if it exists. 
Otherwise, let $\evaluate{g}$ return $\kw{None}$.
Let $\kw{count}(g) = i$ return the current number of times $\evaluate{g}$ has been called.
A {\em conditional generator} $f(\bar{X})$ is a function from an object tuple $\bar{x}$ to a generator $f(\bar{x}) = g_{\bar{x}}$ that produces a sequence of {\em output} object tuples $g_{\bar{x}}$ that relate to {\em input} object tuple $\bar{x}$.

A {\em stream} $s$ is a conditional generator $s(\bar{X})$ endowed with a declarative specification of any facts its inputs and outputs always satisfy. 
Let $\domain{s} = \{p \mid \forall \bar{x} \in \bar{X}.\; p(\bar{x})\}$ be a set of facts $p$ on input parameters $\inp{s}$ that specify the set of object tuples $\bar{x}$ for which $s(\bar{X})$ is defined.
Let $\graph{s} = \{p \mid \forall \bar{x} \in \bar{X}, \forall \bar{y} \in s(\bar{x}).\; p(\bar{x} + \bar{y})\}$ be a set of {\em \cer{}} predicates on both $\inp{s}$ and output parameters $\out{s}$ that assert any facts that $\langle \bar{x}, \bar{y} \rangle$ pairs satisfy.
Intuitively, domain facts specify ``typing" information by declaring legal inputs, and certified facts declare properties that all outputs are guaranteed to satisfy.
A {\em stream instance} $s(\bar{x})$ is a stream $s$ with its input parameters $\inp{s}$ replaced by an object tuple $\bar{x}$.
Let $s(\bar{x}) {\to} \bar{y}$ denote a stream instance $s(\bar{x})$ that generates output object tuple $\bar{y}$.
An {\em external cost function} $c(\bar{X}) {\to} [0, \infty)$ is a nonnegative function defined on parameter tuple $\bar{X}$. 
Like streams, the domain of $c$ is declared by a set of facts $\domain{c}$ on inputs $\bar{X}$.
However, external cost functions do not produce objects or certify facts.

A PDDLStream {\em problem} $({\cal A}, {\cal S}, {\cal I}, {\cal G})$ is given by a set of actions ${\cal A}$, a set of streams ${\cal S}$, an initial state ${\cal I}$, and a goal state set ${\cal G}$.
To ensure PDDLStream is Turing-recognizable, we require that stream-\cer{} predicates are never negated within action preconditions. 
The set of streams ${\cal S}$ augments the initial state ${\cal I}$, recursively defining a potentially infinite set of facts ${\cal I}^*$ that hold initially and cannot be changed:
\begin{align*}
{\cal I}^* &= {\cal I} \cup \{p(\bar{x} + \bar{y}) \mid s \in {\cal S}, |\bar{x}| = |\inp{s}|, \\
&\forall p' \in \domain{s}.\; p'(\bar{x}) \in {\cal I}^*, \bar{y} \in s(\bar{x}), p \in \graph{s}\}.
\end{align*}

A {\em solution} $\pi$ for PDDLStream problem $({\cal A}, {\cal S}, {\cal I}, {\cal G})$ is a plan such that $\proc{preimage}(\pi + [{\cal G}]) \subseteq {\cal I}^*$. 
For cost-sensitive planning, the objective is to minimize the sum of solution action costs. 
In appendix~\ref{sec:appendix},
we prove that PDDLStream planning is {\em undecidable}, but prove our algorithms are {\em semi-complete}, {\it i.e.}, complete over feasible instances.

\subsection{Domain Description}

In order to enable easy use for AI practitioners, PDDLStream adheres to the PDDL standard when possible and adapts PDDL style and syntax when describing streams.
PDDL problems are typically described using text files. 
A \pddlsmall{domain.pddl} file specifies the domain dynamics through a set of actions (\pddlsmall{:action}) and derived predicates (\pddlsmall{:derived}). 
A \pddlsmall{problem.pddl} file specifies the problem instance through a set of objects (\pddlsmall{:objects}), the initial state (\pddlsmall{:init}), and a goal formula (\pddlsmall{:goal}).

In order to represent first-class objects such as real-valued vectors and implement conditional generators that operate on them, PDDLStream problems are partially described using a programming language. 
However, the declarative components of PDDLStream are still described in PDDL.
Actions and derived predicates are listed using a standard \pddlsmall{domain.pddl} text file.
The input parameters (\pddlsmall{:inp}), domain facts (\pddlsmall{:dom}), output parameters (\pddlsmall{:out}), and certified facts (\pddlsmall{:cert}) of each stream are specified in a \pddlsmall{stream.pddl} text file using PDDL-style syntax.

The conditional generator for each stream is stored programmatically in a map from each stream name to its generator function.
Because the initial state typically contains many constant objects that may be non-string entities, the initial state and goal formula are also expressed programmatically instead of using a \pddlsmall{problem.pddl} text file.



\section{Example Domains} \label{sec:example}

We apply PDDLStream to model two robotic manipulation domains with a single manipulator and a finite set of movable blocks.
{\em Domain 1} (figure~\ref{fig:domain2}) is mobile manipulation task requiring a PR2 robot to tightly pack each blue block into the green region.
The goal in {\em Domain 2} (figure~\ref{fig:domain1-plots}) is to place one of the two blue blocks on the green region while minimizing the robot distance traveled.
The right blue block is much closer to the robot and the goal region than the distant left blue block.
However, the red block must be moved out of way in order to safely grasp the right blue block.
Optimal plans, which pick the near blue block, require more actions but travel less distance than plans that pick the far blue block.

Our model uses the following parameters:
\pddlsmall{?b} is the name of a block;
\pddlsmall{?r} is the name of a region on a stable surface;
\pddlsmall{?p} is 6 DOF block pose placed stably on a fixed surface;
\pddlsmall{?g} is a 6 DOF block grasp transform relative to the robot gripper;
\pddlsmall{?q} is an 11 DOF robot configuration; and 
\pddlsmall{?t} is a trajectory composed of a finite sequence of waypoint robot configurations.
The fluent predicates \pddlsmall{AtConf}, \pddlsmall{AtPose}, \pddlsmall{Holding}, \pddlsmall{Empty} model the changing robot configuration, object poses, and gripper status. 
The static predicates \pddlsmall{Block}, \pddlsmall{Conf}, \pddlsmall{Pose}, \pddlsmall{Grasp}, \pddlsmall{Kin}, \pddlsmall{Motion}, \pddlsmall{Contain}, \pddlsmall{CFree} are constant facts. 
\pddlsmall{Block} declares that \pddlsmall{?b} is a block.
\pddlsmall{Conf} declares that \pddlsmall{?q} is a robot configuration.
\pddlsmall{Pose} and \pddlsmall{Grasp} indicate that a pose \pddlsmall{?p} or grasp \pddlsmall{?g} can be used for block \pddlsmall{?b}. 
\pddlsmall{Kin} is a kinematic constraint. 
\pddlsmall{Motion} is a constraint that \pddlsmall{?q1}, \pddlsmall{?q2} are the start and end configurations for trajectory \pddlsmall{?t}, and \pddlsmall{?t} respects joint limits, self-collisions, and collisions with the fixed environment.
\pddlsmall{Contain} states that when block \pddlsmall{?b} is at pose \pddlsmall{?p}, it is within region \pddlsmall{?r}.
\pddlsmall{CFree} states that if block \pddlsmall{?b} were placed at pose \pddlsmall{?p}, the robot, executing trajectory \pddlsmall{?t}, would not collide with it. 
The cost function \pddlsmall{Dist} gives the distance traveled along trajectory \pddlsmall{?t}.
The \pddlsmall{domain.pddl} file is specified as follows:

\begin{footnotesize}
\begin{lstlisting}
(|\textbf{:derived}| (In ?b ?r)
 (|\textbf{exists}| (?p) (|\textbf{and}| (Contain ?b ?p ?r) 
                   (AtPose ?b ?p))))
(|\textbf{:derived}| (Safe ?t ?b) (|\textbf{or}| 
 (|\textbf{exists}| (?g) (|\textbf{and}| (Grasp ?b ?g)
                   (Holding ?b ?g)))
 (|\textbf{exists}| (?p) (|\textbf{and}| (CFree ?t ?b ?p)
                   (AtPose ?b ?p)))))
(|\textbf{:action}| move
 |\textbf{:param}| (?q1 ?t ?q2)
 |\textbf{:pre}| (|\textbf{and}| (Motion ?q1 ?t ?q2) (AtConf ?q1) 
(|\textbf{forall}|(?b)(|\textbf{imply}| (Block ?b) (Safe ?t ?b))))
 |\textbf{:eff}| (|\textbf{and}| (AtConf ?q2) (|\textbf{not}| (AtConf ?q1))
      (|\textbf{incr}| (|\textbf{total-cost}|) (Dist ?t)))
(|\textbf{:action}| pick
 |\textbf{:param}| (?b ?p ?g ?q)
 |\textbf{:pre}| (|\textbf{and}| (Kin ?b ?p ?g ?q) (AtPose ?b ?p) 
           (Empty) (AtConf ?q))
 |\textbf{:eff}| (|\textbf{and}| (Holding ?b ?g)
      (|\textbf{not}| (AtPose ?b ?p)) (|\textbf{not}| (Empty))))
(|\textbf{:action}| place
 |\textbf{:param}| (?b ?p ?g ?q)
 |\textbf{:pre}| (|\textbf{and}| (Kin ?b ?p ?g ?q) (Holding ?b ?g)
           (AtConf ?q))
 |\textbf{:eff}| (|\textbf{and}| (AtPose ?b ?p) (Empty)
      (|\textbf{not}| (Holding ?b ?g))))
\end{lstlisting}
\end{footnotesize}


Three actions are defined: \pddlsmall{move}, \pddlsmall{pick}, and \pddlsmall{place}.
The \pddlsmall{In} derived predicate expresses whether block \pddlsmall{?b} is currently contained within region \pddlsmall{?r} by expressing a condition on its current pose \pddlsmall{?p}.
The \pddlsmall{Safe} derived predicate encodes whether trajectory \pddlsmall{?t} does not collide with placed block \pddlsmall{?b} at its current pose. 
For simplicity, we omit the description of an additional condition within \pddlsmall{move} that checks collisions between grasped blocks and placed blocks. 

The \pddlsmall{stream.pddl} file
is defined below.
The \pddlsmall{poses} stream randomly samples an infinite sequence of stable placements \pddlsmall{?p} for block \pddlsmall{?b} in region \pddlsmall{?r}. 
The \pddlsmall{grasps} stream enumerates a sequence of force-closure grasps \pddlsmall{?g} for block \pddlsmall{?b}. 
The \pddlsmall{ik} stream calls an inverse kinematics solver to sample configurations \pddlsmall{?q} from a 4D manifold of values (due to manipulator redundancy) that enable the robot to manipulate a block \pddlsmall{?b} at pose \pddlsmall{?p} with grasp \pddlsmall{?g}. 
It is important for \pddlsmall{ik} to have \pddlsmall{?p} and \pddlsmall{?g} as input parameters so it can operate on poses and grasp objects in the initial state as well those produced by \pddlsmall{poses} and \pddlsmall{grasps}.
The \pddlsmall{motion} stream repeatedly calls a motion planner to generate safe trajectories \pddlsmall{?t} between pairs of configurations \pddlsmall{?q1}, \pddlsmall{?q2}. 
The \pddlsmall{cfree} stream tests whether block \pddlsmall{?b} when at pose \pddlsmall{?p} is collision free with respect to all robot configurations along trajectory \pddlsmall{?t}.
It is a {\em test stream}, a stream with no output parameters.
If it generates the empty tuple $\langle \; \rangle$, its certified conditions are proven.
As a result, it can be interpreted as a Boolean function.
\pddlsmall{cfree} is checked by calling a collision checker along trajectory \pddlsmall{?t}.
The \pddlsmall{Dist} external cost function returns the sum of the distance between each pair of adjacent configuration waypoints on trajectory \pddlsmall{?t}.

\begin{footnotesize}
\begin{lstlisting}
(|\textbf{:stream}| poses         (|\textbf{:stream}| ik
 |\textbf{:inp}| (?b ?r)           |\textbf{:inp}| (?b ?p ?g)
 |\textbf{:dom}| (|\textbf{and}| (Block ?b)   |\textbf{:dom}| (|\textbf{and}| 
 (Region ?r))           (Pose ?b ?p)
 |\textbf{:out}| (?p)              (Grasp ?b ?g))
 |\textbf{:cert}| (|\textbf{and}| (Pose ?b ?p)|\textbf{:out}| (?q)
 (Contain ?b ?p ?r)))   |\textbf{:cert}| (|\textbf{and}| (Conf ?q) 
(|\textbf{:stream}| grasps         (Kin ?b ?p ?g ?q)))
 |\textbf{:inp}| (?b)             (|\textbf{:stream}| motion
 |\textbf{:dom}| (Block ?b)        |\textbf{:inp}| (?q1 ?q2)
 |\textbf{:out}| (?g)              |\textbf{:dom}| (|\textbf{and}| (Conf ?q1)
 |\textbf{:cert}| (Grasp ?b ?g))   (Conf ?q2))
(|\textbf{:stream}| cfree          |\textbf{:out}| (?t)
 |\textbf{:inp}| (?t ?b ?p)        |\textbf{:cert}| (|\textbf{and}| (Traj ?t) 
 |\textbf{:dom}| (and (Traj ?t)   (Motion ?q1 ?t ?q2)))
 (Pose ?b ?p))         (|\textbf{:function}| (Dist ?t)
 |\textbf{:cert}| (CFree ?t ?b ?p))|\textbf{:dom}| (Traj ?t))
\end{lstlisting}
\end{footnotesize}

\subsection{Rovers Domain}

\begin{figure}[ht]
  \centering
    \includegraphics[width=\columnwidth]{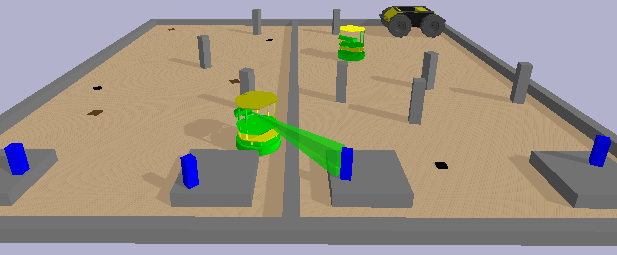}
\caption{{\em Domain 3} (with 4 objectives).} \label{fig:domain1}
\end{figure}

We also apply PDDLStream to a multi-robot surveying domain to demonstrate the generality of our formalism.
{\em Domain 3} (figure~\ref{fig:domain1}) extends the classic PDDL domain {\em rovers}~\cite{long20033rd} by incorporating 3D visibility, distance, reachability, and collision constraints. 
Two rovers (green TurtleBot robots) must together collect a rock sample (black objects), collect a soil sample (brown objects), photograph each objective (blue objects) without occlusions, and communicate the results back to the lander (yellow Husky robot) via line of sight.
Due to obstacles that limit reachability, both rovers must be utilized in order to complete the task.
The actions are: \pddlsmall{move}, \pddlsmall{take\_image},  \pddlsmall{calibrate},  \pddlsmall{send\_image},  \pddlsmall{sample\_rock},  \pddlsmall{send\_analysis},  \pddlsmall{drop\_rock}.

\section{PDDLStream Algorithms} \label{sec:algorithms}\label{sec:level}

We present four PDDLStream algorithms that share several common subroutines.
The first two algorithms ({\em Incremental}, {\em Focused}) are the direct application of the algorithms of Garrett et al.~\shortcite{garrettIJRR2018} to PDDLStream.
The second two algorithms ({\em Binding}, {\em Adaptive}) are new algorithms. 
Each algorithm operates by solving a sequence of finite PDDL problems of increasing size.
Let $\proc{search}({\cal A}, {\cal I}, {\cal G})$ be any sound and complete algorithm for classic PDDL problems.
For cost-sensitive planning, assume $\proc{search}$ returns a solution with cost below a cost bound $C$.
\proc{search} can be implemented using an off-the-shelf PDDL planner without modification to take advantage of existing, efficient search algorithms.
Although each algorithm is presented in its decision form, each can easily be run in an {\em anytime} fashion.


In order to reduce a potentially infinitely-large PDDLStream problem to a sequence of finite PDDL problems, our algorithms control two infinite sources of objects.
First, the generator for a stream instance may enumerate an infinitely large set.
Second, it may be possible to compose a sequence of stream instances of unbounded length. 
Thus, both the maximum {\em width} and {\em depth} of generated objects must be limited. 
We capture both of these properties by introducing the notion of the {\em level} of a fact. 
Intuitively, a level relates to the number of stream evaluations that are required to certify a fact.
The level recursively incorporates both the stream evaluations required to certify its domain facts as well as the number of evaluations of $s(\bar{x})$ itself.
This idea is similar to the concept of layer for facts and actions in a relaxed planning graph~\cite{bonet2001planning} with the distinction that a stream instance can be evaluated many times.

Each algorithm maintains a map $U$ from each certified fact to both the level (\id{level}) of the fact and the stream instance (\id{instance}) that certified it.
More formally, the level of stream instance $s(\bar{x})$ is the maximum level of its domain facts in $U$ plus one more than the count of its past evaluations.
See section~\ref{sec:execution} for an example using levels.

\begin{footnotesize}
\begin{equation*}
\proc{level}(U, s(\bar{x})) = 1 +  \kw{count}(s(\bar{x})) +  \max_{p \in \domain{s}} U[p(\bar{x})].\id{level}
\end{equation*}
\end{footnotesize}

\noindent
To ensure that external cost functions are evaluated on the earliest level possible, define the level of an external cost function instance $c(\bar{x})$ to be the max of its domain, {\it i.e.} $\proc{level}(U, c(\bar{x})) = \max_{p \in \domain{f}} U[p(\bar{x})].\id{level}$.



\section{Incremental Algorithm}  \label{sec:incremental}

The {\em Incremental} algorithm enumerates ${\cal I}^*$ by iteratively increasing the maximum level $l$.
For each level, the subroutine \proc{apply-streams} instantiates and evaluates all stream instances $s(\bar{x})$ at level $k \leq l$ and adds any new certified facts to $U$.
The meta-parameter \proc{output} specifies the procedure that is used to generate output objects when evaluating each stream instances.
In this case, $\proc{output} = \kw{next}$ simply queries the next output tuple in the generator.
Let procedure \proc{instantiate} ground all stream instances that are legal given the input objects in $U$ and the currently certified facts:
\begin{align*}
\proc{instantiate}({\cal S}, U) = \{&s(\bar{x}) \mid\; \forall s \in {\cal S}, \forall p \in \domain{s}. \\
& |\bar{x}| = |\inp{s}|, p(\bar{x}) \in U\}.
\end{align*}

\noindent
The current set of certified facts $U$ becomes the initial state in a PDDL problem $({\cal A}, U, {\cal G})$ that is solved using \proc{search}. 
If \proc{search} finds a plan $\pi$, it is returned as a solution.

\begin{footnotesize}
\begin{codebox}
\Procname{$\proc{incremental}({\cal A}, {\cal S}, {\cal I}, {\cal G}):$}
\li $U = \{f: \langle0, \kw{None}\rangle; f \in {\cal I}\}$ \Comment Map from fact to level 
\li \For $l \in [0, 1, 2, ...]$: \Then
\li $U = \proc{apply-streams}({\cal S}, U, l; \kw{next})$
\li $\pi = \proc{search}({\cal A}, U, {\cal G})$
\li \If $\pi \neq \kw{None}$: \Return $\pi$
\End
\end{codebox}
\begin{codebox}
\Procname{$\proc{apply-streams}({\cal S}, U', l; \proc{output}):$}
\li $U = \kw{copy}(U')$ 
\li \For $k \in [1, 2, ..., l]$: \Then
\li \For $s(\bar{x}) \in \proc{instantiate}({\cal S}, U)$:  \If $\level{U, s(\bar{x})} = k$ \Then
\li $\proc{add-certified}(U, s(\bar{x}); \proc{output})$
\End\End
\li \Return $U$
\end{codebox}
\begin{codebox}
\Procname{$\proc{add-certified}(U, s(x); \proc{output}):$}
\li $l = \level{U, s(\bar{x})}$; $\bar{y} = \proc{output}(\bar{x})$
\li $F = \{p(\bar{x} + \bar{y}) \mid p \in \graph{s}\} \kw{ if } \bar{y} \neq \kw{None} \kw{ else } \emptyset$
\li \For $f \in (F \setminus U)$: $U[f] = \langle l, s(\bar{x}) \rangle$
\li \Return $\bar{y}$
\end{codebox}
\end{footnotesize}

The incremental algorithm {\em eagerly} and blindly evaluates all stream instances, producing many facts that are irrelevant to the task.
This can result in significant overhead when stream evaluations are computationally expensive as they frequently are in robotics domains where inverse kinematics solvers and motion planners are required.


\section{Optimistic Algorithms} \label{sec:focused}


The remaining algorithms ({\em Focused}, {\em Binding}, and {\em Adaptive}) use the shared pseudocode \proc{optimistic}, which takes in a meta-parameter procedure $\proc{process-streams}$ that implements each algorithm.
The key principle behind our algorithms is to {\em lazily} explore candidate plans before checking their validity~\cite{dellin2016unifying}.
In order to apply laziness to PDDLStream, we plan using  {\em optimistic objects} that represent hypothetical stream outputs before evaluating actual stream outputs.
These values are optimistic in the sense that their corresponding stream instance may not ever produce a satisfying value. 
For instance, an \pddlsmall{ik} stream with a particular pose and grasp pair as inputs may not admit any inverse kinematic solutions. 
By first planning with optimistic objects, our algorithms are able to identify only the stream instances that could possibly support a plan and therefore focus sampling on useful aspects of the problem.


Let the procedure $\proc{opt-output}(s(\bar{x})) = \bar{o}_x^s$ create an optimistic object tuple for stream instance $s(\bar{x})$.
Critically, this technique differs from the approach of Garrett et al.~\shortcite{garrettIJRR2018} in that here each optimistic object $\bar{o}_x^s$ is {\em unique} to a single stream instance $s(\bar{x})$.
In contrast, the approach of Garrett et al., if directly applied to PDDLStream, would create an optimistic object tuple $\proc{opt-output}(s(\bar{x})) = \bar{o}^s$ for each {\em stream} rather than each {\em stream instance}.
As a result, $\bar{o}^s$ is {\em shared} among all instances of stream $s$.
This distinction is significant because each {\em unique} optimistic object $\bar{o}_x^s$ implicitly encodes a single partially-ordered set of stream instance evaluations that could produce values for the optimistic object.
This property provides the basis for our novel {\em Binding} (section~\ref{sec:binding}) and {\em Adaptive} (section~\ref{sec:adaptive}) algorithms.

A consequence of creating unique optimistic objects is that the set of all optimistic objects may be infinitely large in domains where it is possible to compose arbitrarily many streams instances.
In contrast, creating shared optimistic objects always results in a finite set of optimistic objects.
In order to limit the number of unique optimistic objects, we regulate the current set of optimistic stream instances using their level (section~\ref{sec:level}).
Namely, we iteratively increase the maximum optimistic stream level $l$ that can be considered on a given iteration.
Finally, when applied to an external cost function instance $c(\bar{x})$, let $\proc{opt-output}(c(\bar{x})) = 0$ produce an optimistic evaluation of $c(\bar{x})$ by returning 0, a lower bound on the nonnegative cost function value.


\begin{footnotesize}
\begin{codebox}
\Procname{$\proc{optimistic}({\cal A}, {\cal S}, {\cal I}, {\cal G}; \proc{process-streams}):$}
\li $U = \{f: \langle0, \kw{None}\rangle \mid f \in {\cal I}\}$ \Comment Map from fact to level 
\li \For $l \in [0, 1, 2, ...]$: \Then
\li \While \kw{True}: \Then
\li $U^* = \proc{apply-streams}({\cal S}, U, l; \proc{opt-output})$
\li $\pi^* = \proc{search}({\cal A}, U^*, {\cal G})$ 
\li \kw{if} $\pi^* = \kw{None}$: \kw{break}
\li $\psi = \proc{retrace}(U, U^*, \proc{preimage}(\pi^* + [{\cal G}]))$
\li $\pi = \proc{process-streams}(U, \psi, \pi^*)$
\li \kw{if} $\pi \neq \kw{None}$: \Return $\pi$
\End\End
\end{codebox}
\begin{codebox}
\Procname{$\proc{retrace}(U, U^*, F):$}
\li $\psi = [\;]$ \Comment Initialize stream plan
\li \For $f \in (F \setminus U)$: \Then
\li $s(\bar{x}) = U^*[f].\id{instance}$
\li $\psi \pluseq \proc{retrace}(U, U^*,  \{p(\bar{x}); p \in \domain{s}\}) {+} [s(\bar{x})]$ 
\End
\li \Return $\psi$
\end{codebox}
\end{footnotesize}


The outer loop of \proc{optimistic} iteratively increases the maximum fact level $l$. 
The inner loop identifies all stream instances at fact level $l$ that optimistically support a plan.
On each iteration of the while loop, \proc{apply-streams} instantiates and optimistically evaluates all stream instances $s(\bar{x})$ at level $k \leq l$, this time using $\proc{output} = \proc{opt-output}$.
This results in $U^*$, a map of all optimistic facts achievable at fact level $l$.
Next, \proc{optimistic} calls \proc{search} to find an optimistic plan $\pi^*$ for the PDDL problem $({\cal A}, U^*, {\cal G})$. 
If $\pi^* = \kw{None}$, no more plans can be found at the current fact level.
And so \proc{optimistic} breaks out of the while loop and increases the fact level to $l+1$. 
Otherwise, \proc{retrace} extracts a {\em stream plan} $\psi$ of stream instances that, presuming successful evaluations, certify the optimistic facts present in the preconditions of $\pi^*$. 
For each optimistic fact in the preimage of $\pi^*$, \proc{retrace} adds the stream instance $s(\bar{x})$ that produced it to $\psi$ and recursively applies \proc{retrace} to the domain facts of $s(\bar{x})$.
Once a stream plan $\psi$ is identified, the meta-parameter procedure $\proc{process-streams}$ evaluates a subsequence of $\psi$ and returns a solution $\pi$ if one is found.

\subsection{Focused Algorithm}

The {\em Focused} algorithm implements \proc{process-streams} using the procedure \proc{focused-process-streams}.
If $\psi = [\:]$, the plan $\pi^*$ uses no optimistic objects and is returned as a solution.
Otherwise, \proc{focused-process-streams} evaluates streams instances that have satisfied domain facts and adds new certified facts to $U$.
The first stream instance $\psi[0]$ is always evaluated.
Because evaluation with \kw{next} increments the level of $s(\bar{x})$, the same stream plan $\psi$ cannot be used on the following iteration.
This forces \proc{search} to find an optimistic plan $\pi^*$ supported by a new stream plan or report that no more exist, causing the level $l$ to increase.


\begin{footnotesize}
\begin{codebox}
\Procname{$\proc{focused-process-streams}(U, \psi, \pi^*):$}
\li \kw{if} $\psi = [\;]$: \Return $\pi^*$
\li \For $s(\bar{x}) \in \psi$: \If $\{p(\bar{x}) \mid p \in \domain{s}\} \subseteq U$: \Then
\li $\proc{add-certified}(U, s(\bar{x}); \kw{next})$
\End\End
\li \Return \kw{None}
\end{codebox}
\end{footnotesize}

\section{Binding and Adaptive Algorithms}

The primary shortcoming of {\em Focused} is that it fails to take full advantage of the plans produced by \proc{search}.
Our two new algorithms implement \proc{process-streams} by operating on more of  the associated stream plans at a time.
Ultimately, our {\em Adaptive} algorithm balances the time spent in \proc{search} versus \proc{process-streams}, often reducing the number of calls to \proc{search} required to find a solution.

\subsection{Binding Algorithm} \label{sec:binding}

The key idea of {\em Binding} is to propagate stream outputs that are inputs to subsequent streams to evaluate more of the stream plan at once.
\proc{process-streams-binding} maintains a set of {\em bindings} $B$, assignments of each optimistic object to an actual object, that are produced while evaluating the stream plan $\psi$.
Bindings are used to replace any optimistic objects that serve as stream instance inputs in $\psi$ or action arguments in $\pi^*$.
Recall from section~\ref{sec:focused} that optimistic objects $o^s_{\bar{x}}$ are {\em unique} to a particular stream instance $s(\bar{x})$.
Thus, there is a bijective mapping between each optimistic object $o^s_{\bar{x}}$ and its corresponding output object from $s(\bar{x})$.
The procedure \proc{update-bindings} substitutes the optimistic objects in $\bar{x}^*$ with their bindings $\bar{x}$ from $B$, evaluates the stream instance $s(\bar{x})$, and if an output tuple $\bar{y} \neq \kw{None}$ is produced, updates $B$ by mapping each optimistic output $y^*$ to its new object $y$.
If all stream evaluations are successful, then $\psi$ is {\em satisfied}, and procedure \proc{apply-bindings} substitutes each optimistic object within $\pi^*$ with its value in $B$ and returns the new plan as a solution. 
If a stream instead returns $\kw{None}$ or the evaluated cost exceeds the current cost bound $C$, \proc{binding-process-streams} terminates early to avoid unnecessarily evaluating any subsequent stream instances.

\begin{footnotesize}
\begin{codebox}
\Procname{$\proc{binding-process-streams}(U, \psi, \pi^*):$}
\li $B = \{\;\}$ \Comment Initialize bindings
\li \For $s(\bar{x}^*) \in \psi$: \Then
\li $B = \proc{update-bindings}(U, B, s(\bar{x}))$
\li \If $B = \kw{None}$: \Return \kw{None}
\End\End
\li \Return \proc{apply-bindings}($B, \pi^*$)
\end{codebox}
\begin{codebox}
\Procname{$\proc{update-bindings}(U, B, s(\bar{x}^*)):$}
\li $\bar{x} = [B[x^*] \kw{ if } x^* \in B \kw{ else } x ^* \kw{ for } x^* \in \bar{x}^*]$
\li $\bar{y} = \proc{add-certified}(U, s(\bar{x}); \kw{next})$
\li \If $\bar{y} = \kw{None}$: \Return \kw{None}
\li \For $y^*, y \in \kw{zip}(\proc{opt-output}(s(\bar{x})), \bar{y})$: $B[y^*] = y$
\End\End
\li \Return $B$
\end{codebox}
\end{footnotesize}

The performance of {\em Binding} depends on the number times \proc{binding-process-streams} fails to bind each stream plan $\psi$ that is considered. 
And the likelihood that \proc{binding-process-streams} fails depends on the properties of the streams specified for a domain, such as the fraction of stream instances that fail to produce output values ($\evaluate{s(\bar{x}}) = \kw{None}$), as well as the objects present in a specific problem instance.
For example, in {\em Domain 1}, the first optimistic plan considered is always satisfiable; however, most calls to \proc{binding-process-streams} fail due to fact that the \pddlsmall{cfree} stream often fails due to the highly-constrained nature of packing blocks into a small region.
In {\em Domain 2}, the first optimistic plan is never satisfiable because the red block obstructs all ways of picking the blue block, but an optimistic plan that first moves the red block and then the blue block admits many bindings.
In {\em Domain 3}, if a rover configuration sampled to photograph a particular objective is not reachable, it is likely that most configurations sampled for that particular rover and objective pair are not reachable.

\subsection{Example Execution} \label{sec:execution}
As an example of \proc{binding-process-streams}, consider a PDDLStream problem in the robotics domain (section~\ref{sec:example}) requiring that block \pddlsmall{b} be moved from initial pose $p_0$ to a goal region \pddlsmall{r}. 
The objects $q_0, p_0, g_1, t_1, ...$ are real-valued vectors ({\it e.g.} $q_0 = [1.71, -2.44, ...]$).
The initial state is:

\begin{footnotesize}
\begin{lstlisting}
${\cal I} = \{$(Region r) (Block b) (Pose b $p_0$)
(Conf $q_0$) (AtPose b $p_0$) (Empty) (AtConf $q_0$)$\}$.
\end{lstlisting}
\end{footnotesize}
\noindent
The goal is {\small ${\cal G} = \{(\pddl{InRegion b r})\}$}.
\proc{optimistic} fails to find a plan for level $l \leq 2$.
When $l=3$, the optimistic stream instances instantiated by \proc{apply-streams} are:
\begin{footnotesize}
\begin{align*}
[&\pddl{grasps}(\pddl{b}) {\to} \pmb{\gamma}_1, \pddl{poses}(\pddl{b}, \pddl{r}) {\to} \pmb{\rho}_1, \pddl{ik}(\pddl{b}, p_0, \pmb{\gamma}_1) {\to} \pmb{\zeta}_1, \\
& \pddl{ik}(\pddl{b}, \pmb{\rho}_1, \pmb{\gamma}_1) {\to} \pmb{\zeta}_2, \pddl{motion}(q_0, q_0) {\to} \pmb{\tau}_1,\\ 
&\pddl{motion}(q_0, \pmb{\zeta}_1) {\to} \pmb{\tau}_2, \pddl{motion}(\pmb{\zeta}_1, q_0) {\to} \pmb{\tau}_3, ...]. 
\end{align*}
\end{footnotesize}
\noindent
Each $\pmb{\gamma}_i, \pmb{\rho}_i, \pmb{\zeta}_i$, and $\pmb{\tau}_i$ represents a unique optimistic output.
In total, 13 stream instances are created.
Here, the \pddlsmall{poses} and \pddlsmall{grasps} stream instances are all level $1$,  the \pddlsmall{ik} stream instances are all level $2$, and the \pddlsmall{motion} stream instances are all level $3$. 
A possible optimistic plan $\pi^*_1$ and stream plan $\psi_1$ produced by \proc{search} and \proc{retrace} are:
\begin{footnotesize}
\begin{align*}
&\pi^*_1 = [ \pddl{move}(q_0, \pmb{\tau}_2, \pmb{\zeta}_1), \pddl{pick}(\pddl{b}, p_0, \pmb{\gamma}_1, \pmb{\zeta}_1), \pddl{move}(\pmb{\zeta}_1, \pmb{\tau}_4, \pmb{\zeta}_2), \\
&\pddl{place}(\pddl{b}, \pmb{\rho}_1, \pmb{\gamma}_1, \pmb{\zeta}_2) ] \\
&\psi _1= [ \pddl{grasps}(\pddl{b}) {\to} \pmb{\gamma}_1, \pddl{poses}(\pddl{b}, \pddl{r}) {\to} \pmb{\rho}_1, \pddl{ik}(\pddl{b}, p_0, \pmb{\gamma}_1) {\to} \pmb{\zeta}_1\\
&\pddl{ik}(\pddl{b}, \pmb{\rho}_1, \pmb{\gamma}_1) {\to} \pmb{\zeta}_2, \pddl{motion}(q_0, \pmb{\zeta}_1) {\to} \pmb{\tau}_2, \pddl{motion}(\pmb{\zeta}_1, \pmb{\zeta}_2) {\to} \pmb{\tau}_4 ] 
\end{align*}
\end{footnotesize}
\noindent
Assuming each stream evaluation is successful, the following objects are produced, which correspond to bindings $B = \{\pmb{\gamma_1}: g_1, \pmb{\rho_1}: p_1,  \pmb{\zeta_1}: q_1, \pmb{\zeta_2}: q_2, \pmb{\tau_2}: t_1, \pmb{\tau_4}: t_2\}$.
After substituting these values for their corresponding optimistic objects in $\pi^*_1$, the plan $\pi_1$ is returned as a solution.

\begin{footnotesize}
\begin{align*} \label{eqn:sequence}
&\evaluate{\pddl{grasps}(\pddl{b})} = g_1, \evaluate{\pddl{poses}(\pddl{b}, \pddl{r})} = p_1\\ 
&\evaluate{\pddl{ik}(\pddl{b}, p_0, g_1)} = q_1, \evaluate{\pddl{ik}(\pddl{b}, p_1, g_1)} = q_2 \\ 
&\evaluate{\pddl{motion}(q_0, q_1))}= t_1, \evaluate{\pddl{motion}(q_1, q_2)} = t_2  
\end{align*}
\end{footnotesize}
In the event that, for example, an inverse kinematic stream evaluation fails, {\it e.g.} {\small $\evaluate{\pddl{ik}(\pddl{b}, p_0, g_1)} = \kw{None}$}, \proc{binding-process-streams} terminates, and the levels of {\small $\pddl{grasps}(\pddl{b})$} and {\small $\pddl{ik}(\pddl{b}, p_0, g_1)$} are incremented to $2$ and $3$.
As a result, both of the following optimistic stream sequences are only possible when maximum level $l \geq 4$, preventing them from being applied again until $l$ is incremented due to \proc{search} failing to find a plan ($\pi^*_i = \kw{None}$).

\begin{footnotesize}
\begin{align*}
1)\;&[\pddl{grasps}(\pddl{b}) {\to} \pmb{\gamma}_1,  \pddl{ik}(\pddl{b}, p_0, \pmb{\gamma}_1) {\to} \pmb{\zeta}_1, \pddl{motion}(q_0, \pmb{\zeta}_1) {\to} \pmb{\tau}_1] \\
2)\;&[\pddl{ik}(\pddl{b}, p_0, g_1) {\to} \pmb{\zeta}_2, \pddl{motion}(q_0, \pmb{\zeta}_2) {\to} \pmb{\tau}_2]
\end{align*}
\end{footnotesize}

\subsection{Adaptive Algorithm}  \label{sec:adaptive} 


The {\em Binding} algorithm will reconsider each previously identified stream plan $\psi$ using \proc{binding-process-streams}. 
However, it may perform many calls to \proc{search}, each of which is expensive, before $\psi$ can be revisted. 
Rather than always {\em explore} new optimistic plans, it may be beneficial to {\em exploit} our current set of optimistic plans by expending more computation to find feasible bindings for them.
Doing so can be advantageous because these plans can be repeatedly processed without any overhead from \proc{search}.
As a result, an algorithm can {\em aggressively} search through the space of bindings to attempt to find a satisfying set as well as {\em locally optimize} for bindings that correspond to low-cost instantiations of the optimistic plan.
However, there may be stream plans that are not satisfiable, such as in {\em Domain 2} and {\em Domain 3}, so an algorithm still may need to explore additional optimistic plans.
This goal of {\em balancing} the {\em exploration-exploitation trade-off}~\cite{robbins1952some} when planning optimistically is the basis for our {\em Adaptive} algorithm.

Instead of evaluating each stream instance only once, \proc{adaptive-process-streams} maintains a queue $Q$ of bindings to repeatedly consider. 
Each entry contains a stream plan $\psi$, an optimistic plan $\pi^*$, bindings $B$, and the next stream plan index $i$ to process.
$Q$ persists across all invocations and thus contains bindings for previously identified entries that can be reattempted indefinitely.
On each invocation, the queue $Q$ is processed until it is either empty or the time elapsed exceeds a timeout parameter $T$.
The best choice of $T$ varies per domain depending on whether it more beneficial to explore (small $T$) or exploit (large $T$).
We maintain a running sum of the time spent by both \proc{search} and \proc{adaptive-process-streams} as $T_s$ and $T_p$ respectively.
This enables us to {\em adaptively} choose $T \leftarrow \max{(0, T_s - T_p)}$, equating the time spent by both procedures and ensuring that neither dominates the total runtime.

\begin{footnotesize}
\begin{codebox}
\Procname{$\proc{adaptive-process-streams}(U, \psi_+, \pi^*_+; T):$}
\li $Q = [\langle \psi_+, \pi^*_+, \{\;\}, 0 \rangle]$ \Comment Initialize queue with empty binding
\li \While $Q \neq [\;]$ \kw{and} \kw{not} \proc{timeout}($T$): \Do
\li $\psi, \pi^*, B, i = \proc{pop}(Q)$
\li \If $i = \kw{len}(\psi)$: \Return $\proc{apply-bindings}(B, \pi^*)$
\li $B' = \proc{update-bindings}(U, \kw{copy}(B), \psi[i])$
\li \If $B' \neq \kw{None}$: $\proc{push}(Q, \langle \psi, \pi^*, B', i+1 \rangle)$
\li $\proc{push}(Q, \langle \psi, \pi^*, B, i \rangle)$ \Comment Return $\langle \psi, \pi^*,B, i \rangle$ to $Q$ 
\End
\li \Return \kw{None}
\end{codebox}
\end{footnotesize}

Additionally, we implement $Q$ as a priority queue that sorts entries by increasing $\kw{count}(\psi[i])$ followed by $\kw{len}(\psi) - i$.
This approach lexicographically prefers evaluating entries with stream instances $s(\bar{x}) = \psi[i]$ that have been evaluated fewer times followed by stream plans where fewer unbound optimistic objects remain.
This strategy applies the {\em optimism in the face of uncertainty}~\cite{sutton2018reinforcement} principle by prioritizing partially-bound stream plans that have been explored less.
Finally, we continue popping entries off of $Q$, despite the fact that the timeout may be exceeded, as long as $\kw{count}(s_i(\bar{x})) = 0$ in order to greedily evaluate stream instances that have yet to be evaluated.

\begin{figure*}[ht!]
  \centering
  \includegraphics[width=0.34\textwidth]{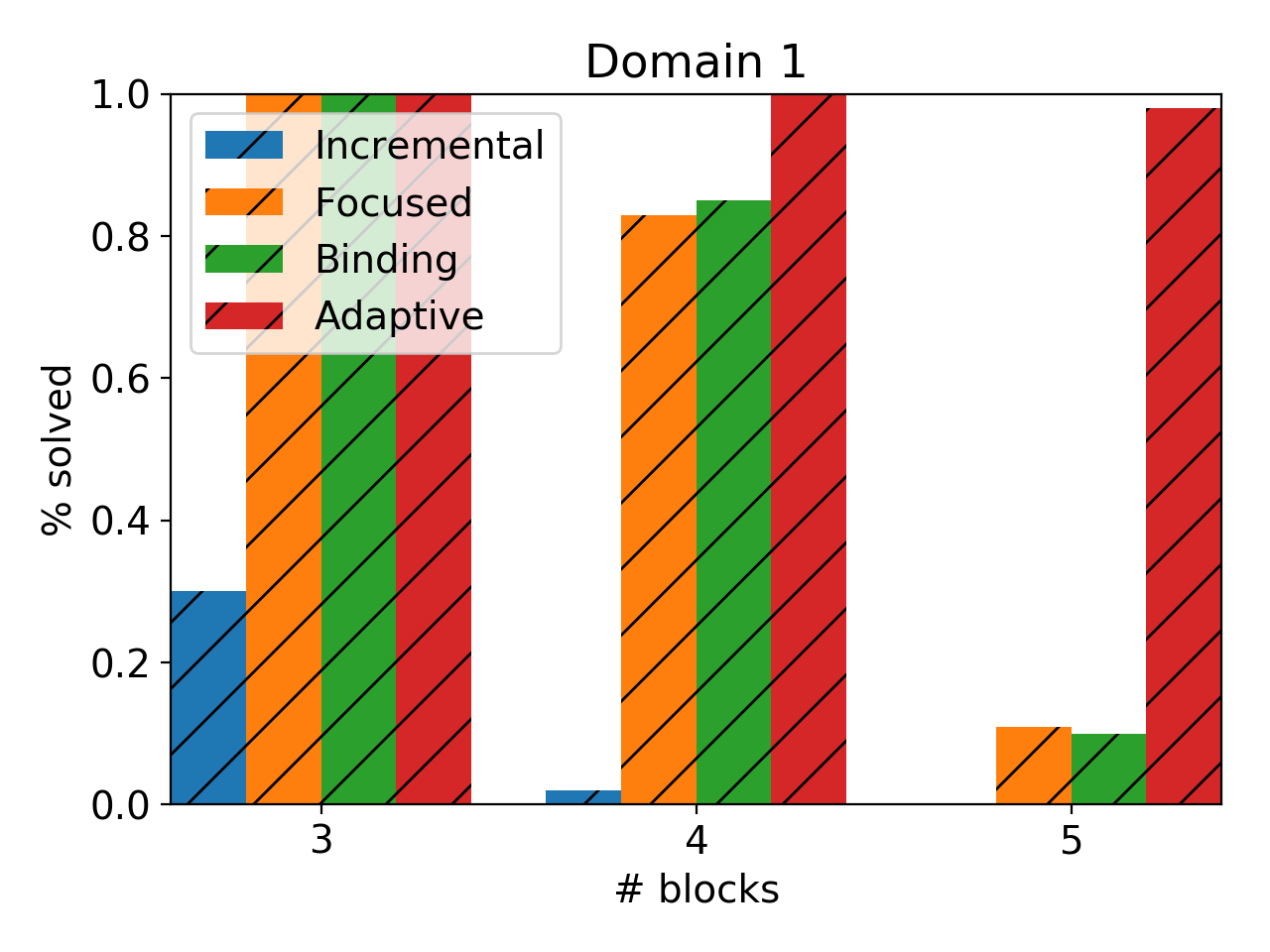}
  \includegraphics[width=0.34\textwidth]{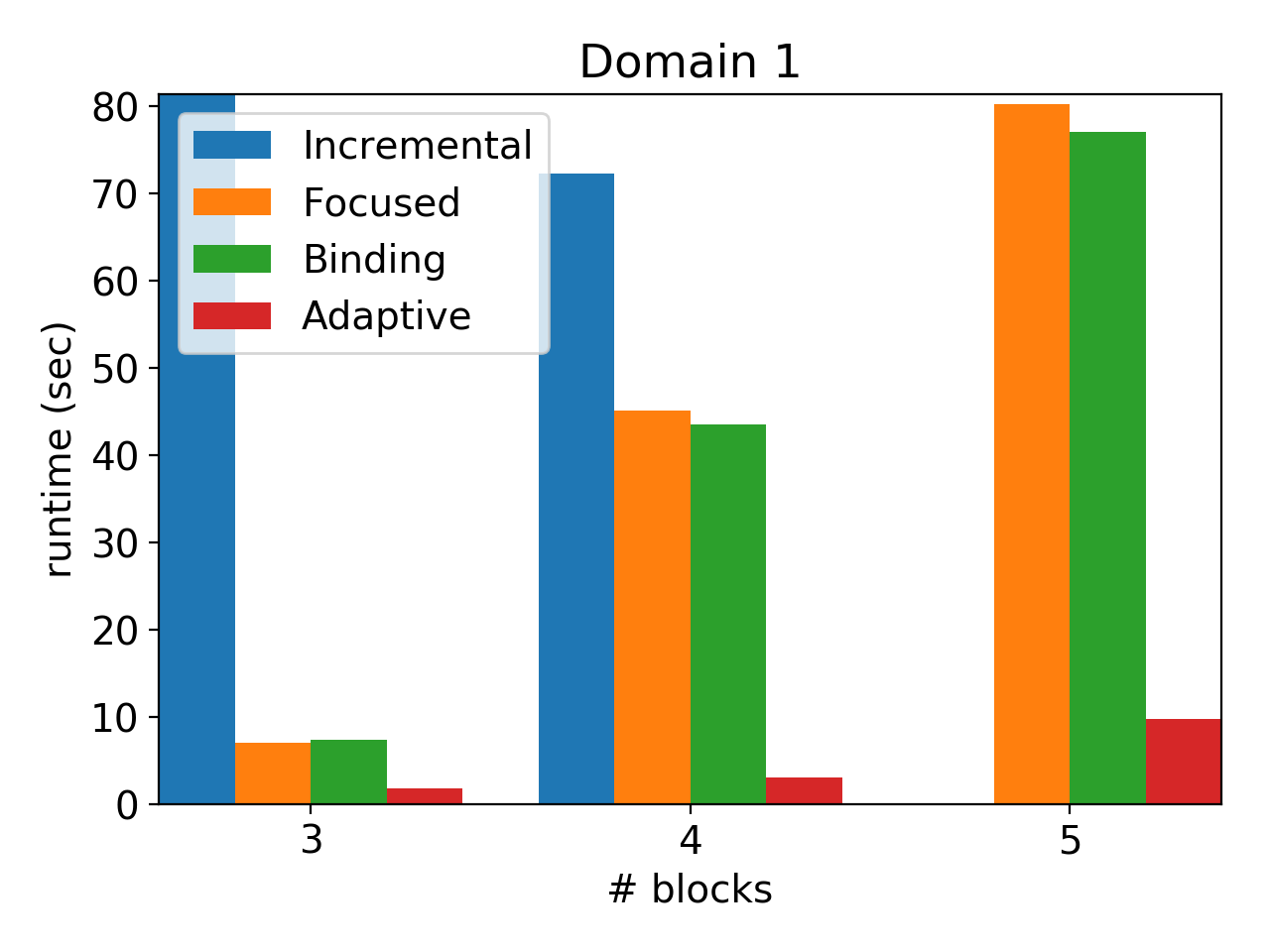}
  \includegraphics[width=0.31\textwidth]{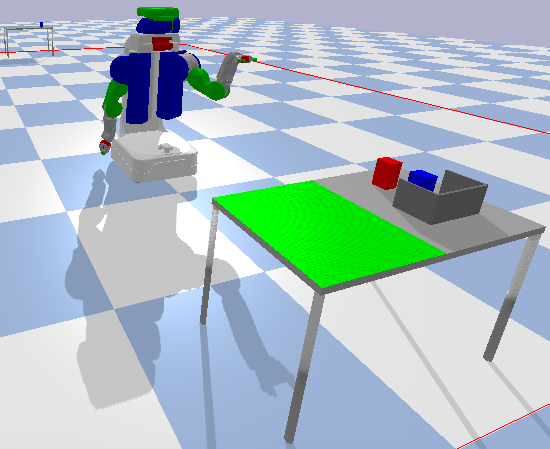}
\caption{From left to right: {\em Domain 1} success percent, {\em Domain 1} mean runtime, and {\em Domain 2}.} \label{fig:domain1-plots}
\end{figure*}

\begin{figure*}[ht!]
  \centering
  \includegraphics[width=0.33\textwidth]{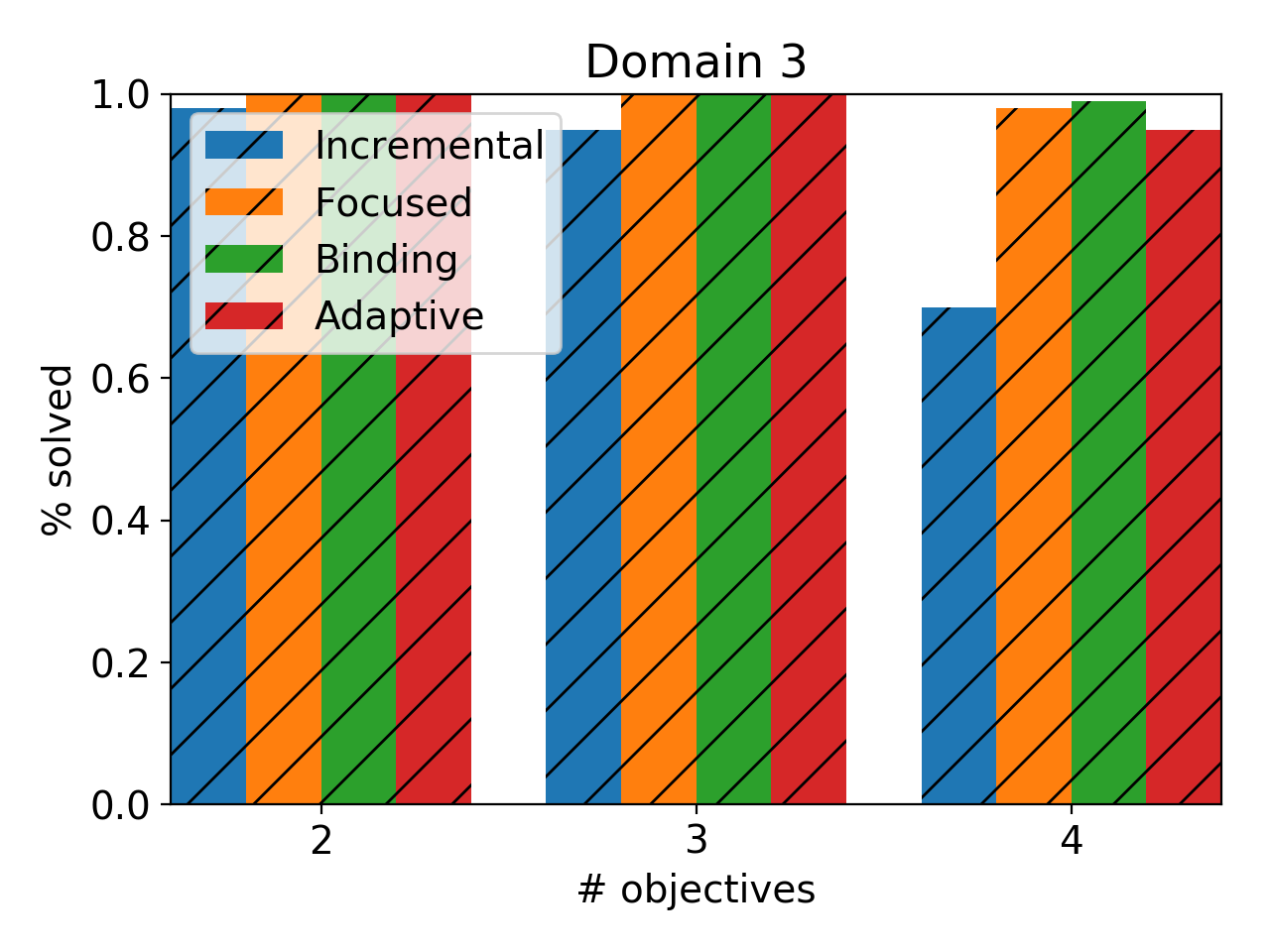}
  \includegraphics[width=0.33\textwidth]{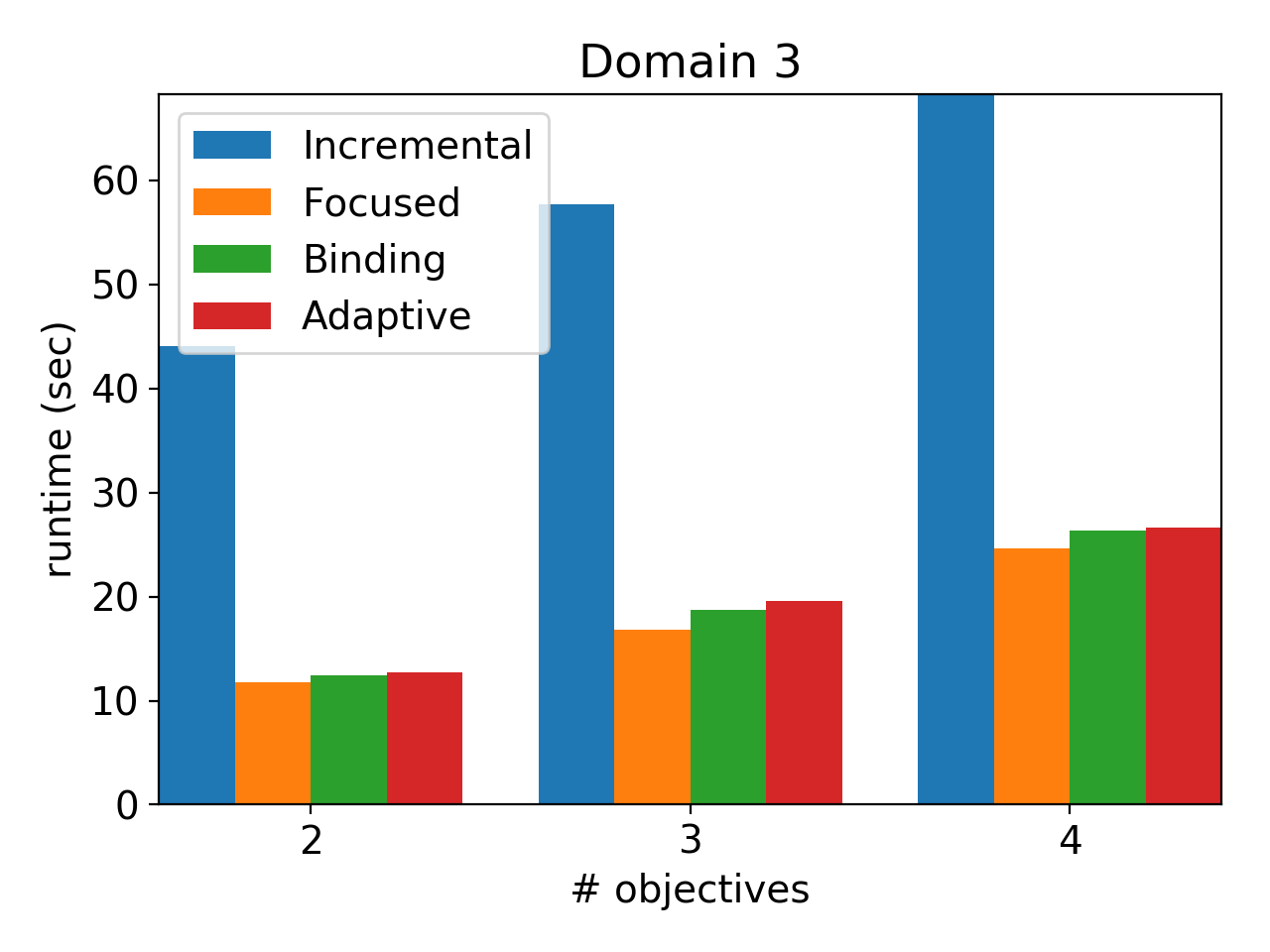}
  \includegraphics[width=0.33\textwidth]{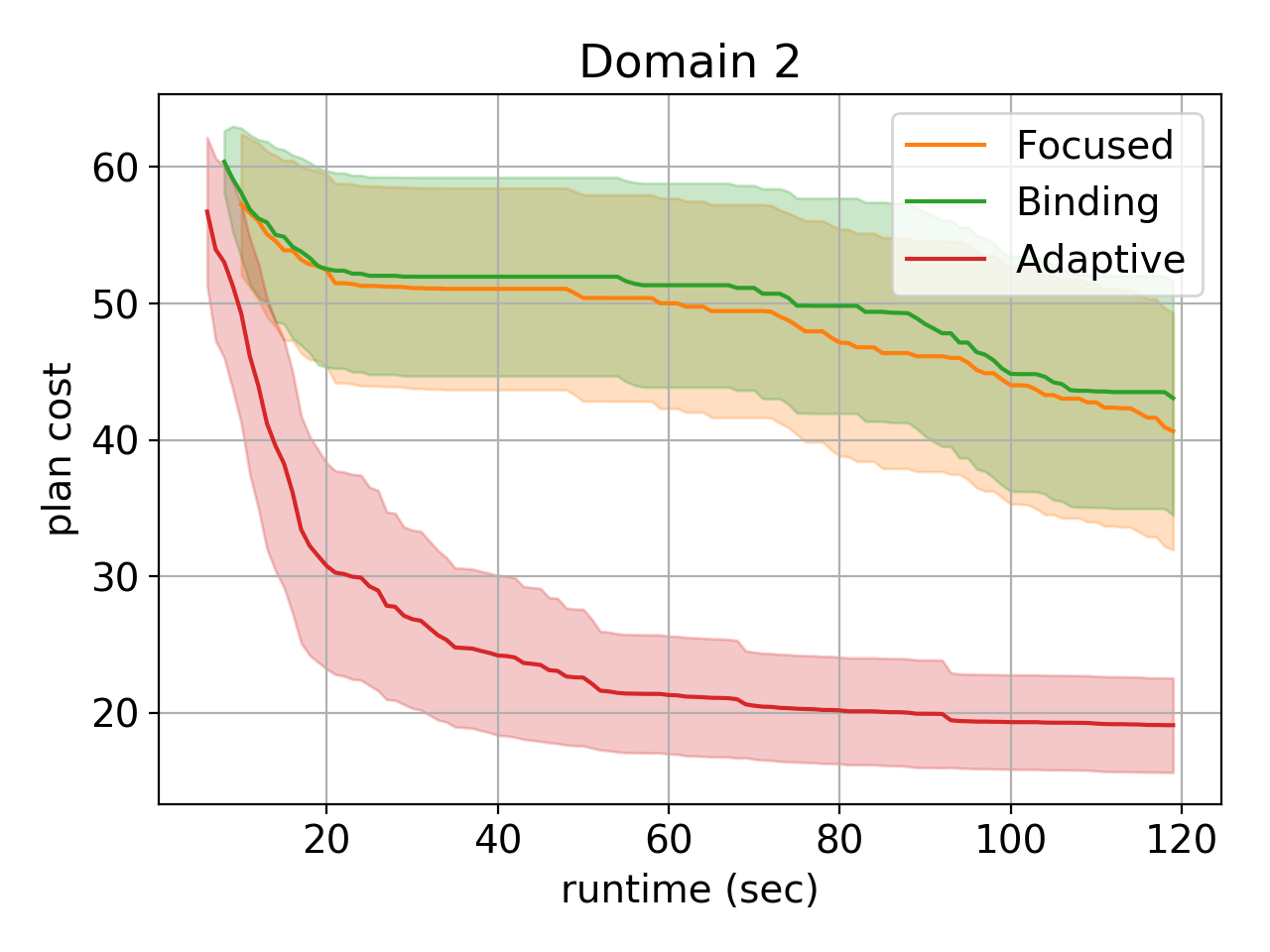}
\caption{From left to right: {\em Domain 3} success percent, {\em Domain 3} mean runtime, and plan cost over time for {\em Domain 2}.} \label{fig:domain2-plots}
\end{figure*}

\subsection{Rebinding}

Optimistic plans may contain objects that were generated by streams.
For example, in {\em Domain 2}, the second optimistic plan $\pi_2^*$ identified (\pddlsmall{move} actions are omitted) has the stream output objects $g_1, q_1, p_1, q_2$ as arguments to the \pddlsmall{pick} and \pddlsmall{place} for the \pddlsmall{blue} block.
Because these objects are not optimistic, they are not present as outputs in stream plan $\psi_2$, and thus \proc{adaptive-process-streams} cannot bind them.

\begin{footnotesize}
\begin{align*}
\pi^*_2 = [&\pddl{pick}(\pddl{red}, p_0', \pmb{\gamma}_1', \pmb{\zeta}_1'), \pddl{place}(\pddl{red}, \pmb{\rho}_1', \pmb{\gamma}_1', \pmb{\zeta}_2'),  \\
&\pddl{pick}(\pddl{blue}, p_0, \underline{g_1}, \underline{q_1}), \pddl{place}(\pddl{blue}, \underline{p_1}, \underline{g_1}, \underline{q_2}) ] 
\end{align*}
\end{footnotesize}
However, the new optimistic objects $\pmb{\gamma}_1', \pmb{\zeta}_1', \pmb{\rho}_1', \pmb{\zeta}_2'$ are still subject to constraints and costs involving the {\em fixed} objects $g_1, q_1, p_1, q_2$. 
For instance, the stream plan tail $[\pddl{motion}(\pmb{\zeta}_2', q_1) {\to} \pmb{\tau}_4, \pddl{Dist}(\pmb{\tau}_4)]$ implicitly tests whether $q_1$ is reachable from $\pmb{\zeta}_2'$ and imposes a cost based on the distance traveled along a trajectory $\tau_4$ between them.
Intuitively, we would instead want to explore combinations of all these arguments as {\em free} parameters.
To do this, we alter line 7 in $\proc{optimistic}$ to be $\psi = \proc{retrace}(\underline{{\cal I}}, U^*, \proc{preimage}(\pi + [{\cal G}]))$, which additionally extracts the sequence of stream instances that produced each {\em non-optimistic} object. 
As a result, fixed objects are now treated as optimistic objects that can take on new values through {\em rebinding}.
This allows \proc{adaptive-process-streams} to explore additional combinations of bindings to more quickly find both feasible and low-cost solutions.

\section{Experiments} \label{sec:results}

We experimented using the {\em Incremental}, {\em Focused}, {\em Binding}, and {\em Adaptive} algorithms on 100 randomly-generated problems within 3 domains in section~\ref{sec:example}. 
The {\em Incremental} and {\em Focused} algorithms serve as baselines that are representative of prior work~\cite{garrettIJRR2018}.
We enforced a 2 minute timeout that includes stream evaluation time.
An open-source Python implementation is available at \url{https://github.com/caelan/pddlstream}.
We use the FastDownward~\cite{helmert2006fast} planning system to implement \proc{search}.
The stream conditional generators were implemented using PyBullet~\cite{coumans2019}.

Figure~\ref{fig:domain2-plots} shows the success rate and mean runtime of successful trials for {\em Domain 1} as the number of blocks increases from 3 to 5, which causes the problem to become more constrained. 
{\em Adaptive} outperforms {\em Incremental}, {\em Focused}, and {\em Binding} due to its ability to aggressively search over many bindings of a single stream plan.
Figure~\ref{fig:domain2-plots} shows the average plan cost over time with a 0.5 standard deviation confidence interval for {\em Domain 2}.
{\em Incremental} is omitted because it only solved 83\% of the problem instances
{\em Adaptive} converges to a low-cost solution more quickly than {\em Focused} and {\em Binding}.
Figure~\ref{fig:domain1-plots} shows the success rate and mean runtime of successful trials for {\em Domain 3} as the number of objectives increases from 2 to 4.
{\em Focused}, {\em Binding}, and {\em Adaptive} all outperform {\em Incremental} and perform about equivalently due to the less geometrically constrained nature of the domain.
The additional stream binding computation only marginally increases the runtime of {\em Adaptive}.

\subsection{Real-World Validation}

We applied PDDLStream to four real-world task and motion planning problems. 
For each task, a PR2 robot observes the initial state, solves for a plan, and executes it in an open-loop fashion.
Our PDDLStream domain description includes 9 actions: \pddlsmall{move}, \pddlsmall{pick}, \pddlsmall{place}, \pddlsmall{stack}, \pddlsmall{push}, \pddlsmall{press}, \pddlsmall{pour}, \pddlsmall{scoop}, \pddlsmall{stir}, and \pddlsmall{cook}. 
Each action is supported by one or more streams that sample its continuous control parameters.
The stream description contains a stream for every manipulation action, each of which samples both the continuous control parameters and ultimately the robot trajectories for executing the associated action.
Figure~\ref{fig:domain2} shows the PR2 solving the {\em serve} task, where it ``prepares a meal" by serving a beverage (blue cup) and a cooked cabbage (green block) on the brown tray.
The robot ``cooks" the cabbage by placing it on the stove, turning the stove on, waiting, and turning the stove off.
Like in  {\em Domain 1}, this problem requires tightly packing the beverage and cabbage on the tray.
{\em Adaptive} is able to quickly identify a collision-free pair of placements supporting a solution.
See appendix~\ref{sec:details}
for descriptions of the other tasks.
Videos of the PR2 completing each task are available at \url{https://tinyurl.com/pddlstream}.




\section{Conclusion}
PDDLStream is a general-purpose framework for incorporating sampling procedures in a planning language.
We introduced two new algorithms that reduce PDDLStream planning to solving a series of finite PDDL problems. 
Our {\em Adaptive} algorithm balances the time spent searching and sampling, allowing it to aggressively explore many possible bindings.
As a result, it outperforms existing algorithms, particularly on tightly-constrained and cost-sensitive problems by greedily optimizing discovered plans.
Finally, we demonstrated that PDDLStream can be used to plan for real-world robots operating using a diverse set of actions.






\bibliographystyle{aaai}
\bibliography{references}

\newpage

\appendix

\section{Theoretical Results} \label{sec:appendix}

Unsurprisingly, PDDLStream planning is undecidable (theorem~\ref{thm:undecidable}) when conditional generators are Turing complete.

\begin{thm} \label{thm:undecidable}
PDDLStream plan existence is undecidable.
\begin{proof}
Consider a trivial reduction from the halting problem.
Given a Turing machine TM, construct a PDDLStream problem with a single stream \pddl{simulate} with no input or output parameters.

\begin{footnotesize}
\begin{lstlisting}
(|\textbf{:stream}| simulate
 |\textbf{:inp}| ()
 |\textbf{:dom}| ()
 |\textbf{:out}| ()
 |\textbf{:cert}| (Reachable))
\end{lstlisting}
\end{footnotesize}

The test stream \pddl{simulate} enumerates the states of TM by simulating one step of TM upon each evaluation.
On an accept state, $\evaluate{\pddl{simulate}()} = ()$, which certifies $\pddl{(Reachable)}$.
Otherwise, $\evaluate{\pddl{simulate}())} = \kw{None}$.
Finally, let ${\cal A} = {\cal I} = \emptyset$ and ${\cal G} = \{\pddl{(Reachable)}\}$.
This problem has a solution if and only if TM halts.
Thus, PDDLStream is undecidable.
\end{proof}
\end{thm}

Although PDDLStream planning is undecidable, it is semi-decidable (implied by theorems~\ref{thm:incremental} and~\ref{thm:focused}).
Thus, we restrict our attention to {\em feasible} problem instances admitting a solution.
We are interested in {\em semi-complete} algorithms for PDDLStream problems, algorithms that are complete over the set of feasible problems.
We will assume \proc{search} can be {\em any} sound and complete PDDL planner. 


In cost-sensitive planning, each action $a$ may have an effect that increases the total plan cost by $c(\bar{X})$.
Because functions may be defined on infinitely large domains, the set of solution costs may not have a minimum.
However, because solution costs are bounded below by zero, this set will have an {\em infimum}.
For example, the sampled set of trajectories between two configurations might converge in cost to a lower bound without actually reaching it.
Because of this, we will only consider the feasibility problem of producing a solution with cost below a specified cost threshold $C$.
This problem is also undecidable but semi-decidable.
In practice, our algorithms can instead be run in an {\em anytime} manner for a bounded amount of time and return the lowest-cost solution identified.
For cost-sensitive planning, we assume $\proc{search}$ is an PDDL planner that returns a solution satisfying cost bound $C$ if a solution exists.


\subsection{Semi-Completeness}

\begin{thm} \label{thm:incremental}
The incremental algorithm is semi-complete.
\begin{proof}
For any solution $\tilde{\pi}$, there exists a finite subset of facts $\tilde{U} = \proc{preimage}(\tilde{\pi} + [{\cal G}])$ in the expanded initial state $\tilde{U} \subseteq {\cal I}^*$ supporting $\tilde{\pi}$.
The incremental algorithm iteratively constructs ${\cal I}^*$ using the set of facts $U$.
After a finite number of iterations $\tilde{U} \subseteq U$.
At which point, $\tilde{\pi}$ will be a solution to the induced PDDL problem.
Because \proc{search} is sound and complete, it will return some solution if not $\tilde{\pi}$ itself.
\end{proof}
\end{thm}

\subsection{Stream Output Uniqueness}

As presented in the main paper, the {\em Focused}, {\em Binding}, and {\em Adaptive} algorithms are all semi-complete under the assumption that each stream output object $y$ is unique to a single stream instance $s(\bar{x})$.
Otherwise, each algorithm requires a modification to preserve semi-completeness.
This discussion is omitted from the main paper for simplicity.

The uniqueness assumption typically holds in practice as streams are often used sample from uncountably infinite sets, such as a bounded interval on the real line.
Consider two streams that each sample independently, uniformly at random from this interval.
The probability that the same value is ever generated by both streams is zero.
Because streams are a component of a PDDLStream problem description, these stochastic streams induce a distribution over PDDLStream problems that arise from the outcomes of each stream.
Thus, the probability of generating a PDDLStream problem that does not have unique output objects is zero.

Without this assumption, an object can have facts certified through being {\em output} (as opposed to an input) of multiple streams.
Consider an object $y$ that is an output of two stream instances $\pddl{s1}()$ and $\pddl{s2}()$, {\it i.e.} 
$\langle y \rangle \in \pddl{s1}()$ and $\langle y \rangle \in \pddl{s2}()$.


\begin{footnotesize}
\begin{multicols}{2}
\begin{lstlisting}
(|\textbf{:stream}| s1
 |\textbf{:inp}| ()
 |\textbf{:dom}| ()
 |\textbf{:out}| (?y)
 |\textbf{:cert}| (P1 ?y) 
 \end{lstlisting}
\begin{lstlisting}
(|\textbf{:stream}| s2
 |\textbf{:inp}| ()
 |\textbf{:dom}| ()
 |\textbf{:out}| (?y)
 |\textbf{:cert}| (P2 ?y) 
\end{lstlisting}
\end{multicols}
\end{footnotesize}

Suppose \proc{optimistic} has already produced $y$ from $\pddl{s1}()$ via $\evaluate{\pddl{s1}()} = \langle y \rangle$, certifying \pddl{(P1 $y$)}.
However, suppose both \pddl{(P1 $y$)} and \pddl{(P2 $y$)} are required to support a solution.
\proc{optimistic} would then need to repeatedly evaluate $\pddl{s2}()$ with the {\em intention} of obtaining $y$ again to certify \pddl{(P2 $y$)}.
This could have been avoided if instead stream \pddl{s2} had $\pddl{?y}$ as an input parameter (\pddl{:inp}) rather than an output parameter (\pddl{:out}).
In which case, \pddl{s2} could certify \pddl{(P2 $y$)} via a membership test rather than through enumeration.

To obtain semi-completeness in domains where not all output objects are unique, the {\em Focused}, {\em Binding}, and {\em Adaptive} algorithms must be modified to automatically create an additional test stream for each stream.
Each test stream reflects the possibility that any existing object may optimistically be the output of the stream.
Consider the stream \pddl{example} and its automatically synthesized test stream \pddl{example-test}.

\begin{footnotesize}
\begin{multicols}{2}
\begin{lstlisting}
(|\textbf{:stream}| example
 |\textbf{:inp}| (?x)
 |\textbf{:dom}| (P3 ?x)
 |\textbf{:out}| (?y)
 |\textbf{:cert}| (P4 ?x ?y) 
 \end{lstlisting}
\begin{lstlisting}
(|\textbf{:stream}| example-test
 |\textbf{:inp}| (?x ?y)
 |\textbf{:dom}| (P3 ?x)
 |\textbf{:out}| ()
 |\textbf{:cert}| (P4 ?x ?y) 
\end{lstlisting}
\end{multicols}
\end{footnotesize}

\noindent
The generator function for $\pddl{example}(\bar{x}) = g_{\bar{x}}$ becomes $\pddl{example-test}(\bar{x} + \bar{y}) = [\langle \; \rangle] \kw{ if } \bar{y} \in g_{\bar{x}} \kw{ else } [\;]$, which returns the empty tuple only if $\bar{y} \in g_{\bar{x}}$.
Because the generator $g_{\bar{x}}$ may be infinitely long, $[\bar{y} \in g_{\bar{x}}] = \kw{True}$ can be identified by incrementally checking whether the $i$th output of $\kw{next}(g_{\bar{x}})$ is equal to $\bar{y}$ where $i = \kw{count}(g_{\bar{x}}) + 1$.

In practice, evaluating $\pddl{example-test}$ is expensive because it requires enumerating the full generator $g_{\bar{x}}$ with the hope of producing $\bar{y}$.
Moreover, as previously described, it is typically extremely unlikely that $\bar{y} \in g_{\bar{x}}$ for an arbitrary object tuple $\bar{y}$.
To prevent these test streams from worsening the performance of the {\em Focused}, {\em Binding}, and {\em Adaptive} algorithms, the test streams can be reserved until the level $l$ exceeds a sufficiently large constant.
Doing so prevents the {\em Focused}, {\em Binding}, and {\em Adaptive} algorithms from optimistically planning with these test streams until they have exhausted many other alternatives.
We now prove that the {\em Focused}, {\em Binding}, and {\em Adaptive} algorithms are semi-complete given this modification.

\begin{thm} \label{thm:focused}
The focused, binding, and adaptive algorithms are semi-complete.
\begin{proof}
For any solution $\tilde{\pi}$, there exists a finite stream plan $\tilde{\psi}$ that, upon evaluation, produces $\tilde{U} \subseteq {\cal I}^*$ supporting $\tilde{\pi}$.
This sequence $\tilde{\psi}$ may include the same stream instance multiple times in the event that multiple evaluations using \kw{next} are required to produce a particular output object.
Let $\tilde{l} = \max_{f \in \tilde{U}} \big(\tilde{U}[f].\id{level}\big) \leq |\tilde{\psi}|$ be the minimum level required to certify all facts in $\tilde{U}$ using stream plan $\tilde{\psi}$.

Now suppose $\tilde{\pi}$ is a solution with the smallest possible level $\tilde{l}$.
For levels $l < \tilde{l}$, \proc{optimistic} will perform a finite number of {\em iterations}, calls to \proc{search}, before failing to finding a plan and breaking out of its inner loop.
The set of stream instances that can be instantiated cannot grow throughout the level.
Each iteration requires a finite number of time steps because \proc{search} is assumed to be complete.
On each iteration where $\pi^* \neq \kw{None}$, at least the first stream instance $\psi[0]$ will be evaluated through $\kw{next}(\psi[0])$.
Any additional evaluations by binding and adaptive can only expedite the number of iterations required.
This prevents \proc{search} from identifying a plan $\pi^*$ that is only supported by the same stream plan $\psi$ on the remaining iterations at this level.
Moreover, this causes the set of stream instances that can be instantiated to strictly shrink. 
Eventually, \proc{search} will fail, returning $\pi^* = \kw{None}$ if only because no more stream instances can be instantiated. 

For level $l = \tilde{l}$, \proc{optimistic} will repeat the same process.
This time, if no more stream instances can be instantiated, then all of $\psi$ must have been evaluated.
Thus, $\tilde{U} \subseteq U$ and \proc{search} will return some solution $\pi'$ if not $\tilde{\pi}$ itself.

\end{proof}
\end{thm}

\section{Experimental Details}  \label{sec:details}

See \url{https://tinyurl.com/pddlstream} for videos demonstrating solutions for each task.

\subsection{Simulated Experiments}

\subsubsection{Domain 1}
The PR2 robot must stow each blue block into the green region. 
The number of blue blocks varies from 3 to 5.
Because the green region remains constant, the problem quickly becomes challenging as the number of blocks increases.
Each block admits two top grasps. 
The initial block poses are sampled uniformly at random on the table subject to collision constraints.

\subsubsection{Domain 2}
The PR2 robot must place one of the two blue blocks in the green region while also minimizing the robot base distance traveled.
The left blue block is far away and thus requires two costly base movements to reach it.
The right blue block is nearby but is obstructed by a movable red block, which would first need to be relocated.
Each block admits four side grasps but no top grasps.

\subsubsection{Domain 3}
The team of two rovers (green TurtleBot robots) must photograph each blue objective.
The number of objectives varies from 2 to 4.
The green view cone indicates a rover is calibrating its camera.
The red view cone indicates a rover is taking a photograph.
Additionally, the rovers must analyze one rock (black square) and one soil (brown square) sample.
Rovers must communicate the images and analyzed samples back to the lander (yellow Husky robot) via the blue ray.
Finally, each rover must return to its initial configuration and empty its sample storage space.
The left rover can only reach configurations in the left component and the right rover can only reach configurations in the right component.
Randomly placed grey obstacles obstruct both visibility and motion.

\subsection{Real-World Experiments}

The {\em serve} task is described in the paper.
For the {\em cook} task, the robot is tasked with cooking the cabbage (green block) by placing it on the stove, turning the stove on, and moving the now cooked cabbage to the brown tray.
Because the robot cannot initially reach the green block with its left arm, our planners must intentionally place the cabbage in an intermediate location to hand it off from the robot's right arm to its left arm.
For the {\em stack} task, the goal is for the green block to be on the purple block and for the blue cup to be on the green block.
Solutions require stacking the three blocks while avoiding destabilizing or knocking over the tower.
For the {\em coffee} task, the robot prepares coffee by pouring coffee from the blue cup into the white bowl, scooping sugar from the brown bowl using its spoon, dumping the sugar into the white bowl, and finally stirring the beverage using its spoon.
In order to avoid spillage, the robot's gripper is subject to an orientation constraint when the robot is holding a cup or spoon that contains coffee or sugar.

\end{document}